\documentclass[letterpaper, 10 pt, conference]{ieeeconf}  

\IEEEoverridecommandlockouts                              

\usepackage[utf8]{inputenc}

\usepackage{url}

\usepackage{graphics} 
\usepackage{amsmath} 
\usepackage{amssymb}  

\usepackage{xcolor}
\usepackage{booktabs}
\usepackage{makecell} 
\usepackage{listings}
\usepackage[frozencache]{minted} 
\usepackage{lmodern}
\usepackage{framed}

\usepackage{tikz}
\usetikzlibrary{positioning,arrows.meta,fit,calc}

\usepackage[color={red!100!green!33},textsize=small]{todonotes}

\newcommand \Until      {\mathbin{\mathcal{U}\kern-.1em}}
\newcommand \Since      {\mathbin{\mathcal{S}\kern-.08em}}

\newcommand \Eventually {\lozenge}

\newcommand{\fix}[2][]{\todo[inline,#1]{#2}}

\newcommand{\var}[1]{\texttt{\detokenize{#1}}}

\usepackage{short_commands}

\title{Sequence Planner - Automated Planning and Control for ROS2-based \\ Collaborative and Intelligent Automation Systems}

\author{Martin Dahl$^{1}$, Endre Erös$^{1}$, Atieh Hanna$^{2}$, Kristofer Bengtsson$^{1}$, Petter Falkman$^{1}$
\thanks{*This work has been supported by UNIFICATION, Vinnova, Produktion 2030 and UNICORN, Vinnova, Effektiva och uppkopplade transportsystem.}
\thanks{$^{1}$M. Dahl, E. Erös, K. Bengtsson, and P. Falkman, Department of Electrical Engineering, Chalmers University of Technology, 412 96 Göteborg, Sweden. {\tt\footnotesize (martin.dahl|endree|kristofer.bengtsson|petter.falkman) @chalmers.se}}%
\thanks{$^{2}$A. Hanna, Group Trucks Operation, Research \& Technology Development (R\&TD), Gropegårdsgatan 2, 40508 Göteborg, Sweden. {\tt\footnotesize atieh.hanna@volvo.com}}%
}

\begin{document}

\maketitle
\thispagestyle{empty}
\pagestyle{empty}

\begin{abstract}
Systems based on the Robot Operating System (ROS) are easy to extend with new on-line algorithms and devices. However, there is relatively little support for coordinating a large number of heterogeneous sub-systems. In this paper we propose an architecture to model and control collaborative and intelligent automation systems in a hierarchical fashion.
\end{abstract}


\section{Introduction}
\label{sec:introduction}
Robotics in production is an increasingly complex field. Off-line and manual programming of specific tasks are today replaced by online algorithms that dynamically performs tasks based on the state of the environment~\cite{akl:rpirw:2016, s16030335}. The complexity will be pushed even further when collaborative robots~\cite{doi:10.1142/S0219843608001303} together with other intelligent and autonomous machines and human operators, will replace more traditional automation solutions. To benefit from these collaborative and intelligent automation systems, also the control systems need to be more intelligent, \emph{reacting to} and \emph{anticipating} what the environment and each sub system will do. Combined with the traditional challenges of automation software development, such as safety, reliability, and efficiency, a completely new type of control system is required.

In order to ease integration and development of different types of online algorithms for sensing, planning, and control of the hardware, various platforms have emerged as middle-ware solutions, one of which stands out is the Robot Operating System (ROS)~\cite{ros}. ROS has been incredibly successful having over 16 million downloads in 2018 alone~\cite{ros2018}. The next generation, ROS2~\cite{ros2}, is currently developed, where the communication architecture is based on the Data Distribution Service (DDS)~\cite{1203555} to enable large scale distributed control architectures. This improvement will pave the way for the use of ROS2-based architectures in real-world industrial automation systems, as will be presented in this paper.

However, enabling integration and communication, while greatly beneficial, is just one part of the challenge. The overall control architecture also needs to plan and coordinate all actions of robots, humans and other devices as well as keeping track of a large amount of state related to them. This has led to several frameworks for composing and executing robot tasks (or algorithms), for example the framework ROSPlan~\cite{Cashmore:2015:RPR:3038662.3038708} that uses PDDL-based models for automated task planning and dispatching, SkiROS~\cite{Rovida2017} that simplifies the planning with the use of a skill-based ontology, eTaSL/eTC~\cite{6942760} that defines a constraint-based task specification language for both discrete and continuous control tasks or CoSTAR~\cite{7989070} that uses Behavior Trees for defining the tasks. However, these frameworks are mainly focused on single robot or single agent applications and lacks features to control large scale, real world collaborative and intelligent industrial automation systems. This paper therefore introduces the control architecture \emph{Sequence Planner} (SP) that can handle these types of systems and that utilizes the power of ROS2.

In Section~\ref{sec:use-case}, a collaborative and intelligent systems use case is introduced, which is used as a running example throughout the paper. Section~\ref{sec:prelims} describe the discrete modeling formalism and notation used. Section~\ref{sec:arch} gives an overview of the proposed architecture, which is expanded upon in Sections~\ref{sec:layer0},~\ref{sec:layer1}, and~\ref{sec:layer2}. The implementation of the architecture is discussed in~\ref{sec:implementation}. Finally, Section~\ref{sec:conclusion} contains some concluding remarks.


\section{The collaborative and intelligent automation systems use case}
\label{sec:use-case}
\begin{figure}[ht]
  \centering
  \begin{tikzpicture}
    \node[rectangle,draw,fill=black,anchor=south west,inner sep=0.2mm] (image) at (0,0) {\includegraphics[width=0.48\textwidth]{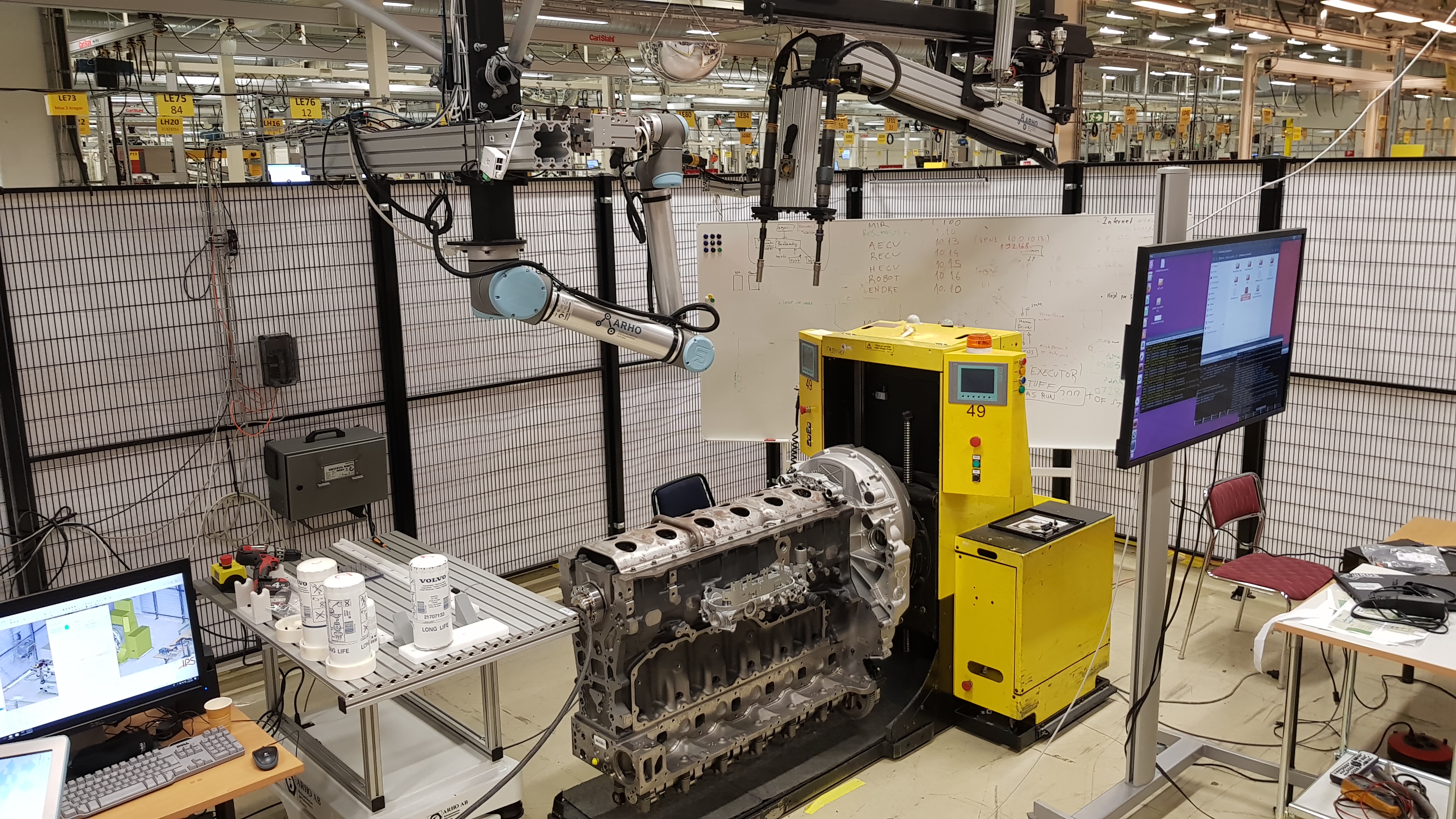}};
  \end{tikzpicture}
  \caption{Collaborative robot assembly station controlled by a network of ROS2 nodes. A video clip from the use case: \protect\url{https://youtu.be/YLZzBfY7pbA}}
  \label{fig:station}
\end{figure}

This paper concerns the development of a ROS2 based automation system for an assembly station in a truck engine manufacturing facility. The challenge involves a collaborative robot and a human operator performing assembly operations on a diesel engine in a collaborative or coactive fashion. In order to achieve this, a wide variety of hardware as well an extensive library of software including intelligent algorithms has to be used. The assembly system can be seen in Figure~\ref{fig:station}.
The physical setup consists of a collaborative robot from Universal Robots, an autonomous mobile platform (a Mir 100), two different specialized end-effectors, a smart nutrunner that can be used by both the robot and the operator, a docking station for the end-effectors, a lifting system for the nutrunner, a camera and RFID reader system and eight computers dedicated for different tasks. The system is communicating over ROS2, with a number of nodes having their own dedicated ROS1 master behind a bridge~\cite{endreFaim}.

The envisioned intelligent and collaborative systems of the future will comprise of several robots, machines, smart tools, human-machine interfaces, cameras, safety sensors, etc. From our experience with this use case, distributed large-scale automation systems require a communication architecture that enable reliable messaging, well-defined communication, good monitoring, and robust task planning and discrete control. ROS1 systems are hard to scale due to a communication layer that wasn't intended for large scale automation use-cases.
However, with ROS2 instead basing its communication on DDS, which has proven real world usage and performance~\cite{6754976}, it seems likely that ROS2 can enable implementation of large scale of industrial automation use-cases.

Systems like this combine the challenges of high level intelligent task and motion planning with the challenges of more traditional automation systems. The automation system needs to keep track of the state of all resources and products, as well as the environment. The control system also needs means of \emph{restarting} production should something go wrong. High level task plans are not sufficient to deal with these complexities, we argue instead that the problem needs to be tackled at the high (task planning) and low (I/O) level simultaneously. Thus the frameworks mentioned in Section~\ref{sec:introduction} are not suitable for use with our use case of the large scale collaborative and intelligent automation systems.

\section{Preliminaries}
\label{sec:prelims}
In this section some background to modeling in Sequence Planner (SP) is briefly described. SP as a modeling tool uses a formal representation of an automation system based on  \emph{extended finite automaton} (EFA)~\cite{skoldstam2007modeling}, a generalization of an automaton that includes guards and actions associated with the transitions. The guards are predicates over a set of variables, which can be updated by the actions. EFA:s allow a reasonably compact representation that is straight forward to translate to problems for different solvers. One example of this is generating bounded model checking problems, used both for falsification and on-line planning in this paper, but also for applying formal verification~\cite{ BF:DASOP:2012} and performing cycle time optimization~\cite{ SWFL:OOSCP:2012}.

\subsection{Extended finite automata}
An extended finite automaton is a 6-tuple: $E=\langle Q \times V,\Sigma,\mc{G},\mc{A}, \ra, (q_0,v_0) \rangle$. The set $Q\times V$ is the extended finite set of states, where $Q$ is a finite set of {\em locations} and $V$ is the finite domain of definition of the variables, $\Sigma$ (the alphabet) is a nonempty finite set of events, $\mc{G}$ is a set of guard predicates over the variables, $\mc{A}$ is a collection of action functions, $\ra \subseteq Q\times \Sigma \times \mc{G}\times \mc{A}\times Q$ is the state transition relation, and $(q_0,v_0)\in Q\times V$ is the initial state. A transition in an EFA is enabled if and only if its corresponding guard formula evaluates to true; when the transition is taken, a set of variables is updated by the action functions. In this work, a transition with a corresponding guard formula and action function is denoted $e : c / a$, where $e \in \Sigma$, $c \in \mc{G}$, and $a \in \mc{A}$.


\section{Architectural overview}
\label{sec:arch}
\begin{figure}[ht]
  \centering
  \includegraphics[width=0.45\textwidth]{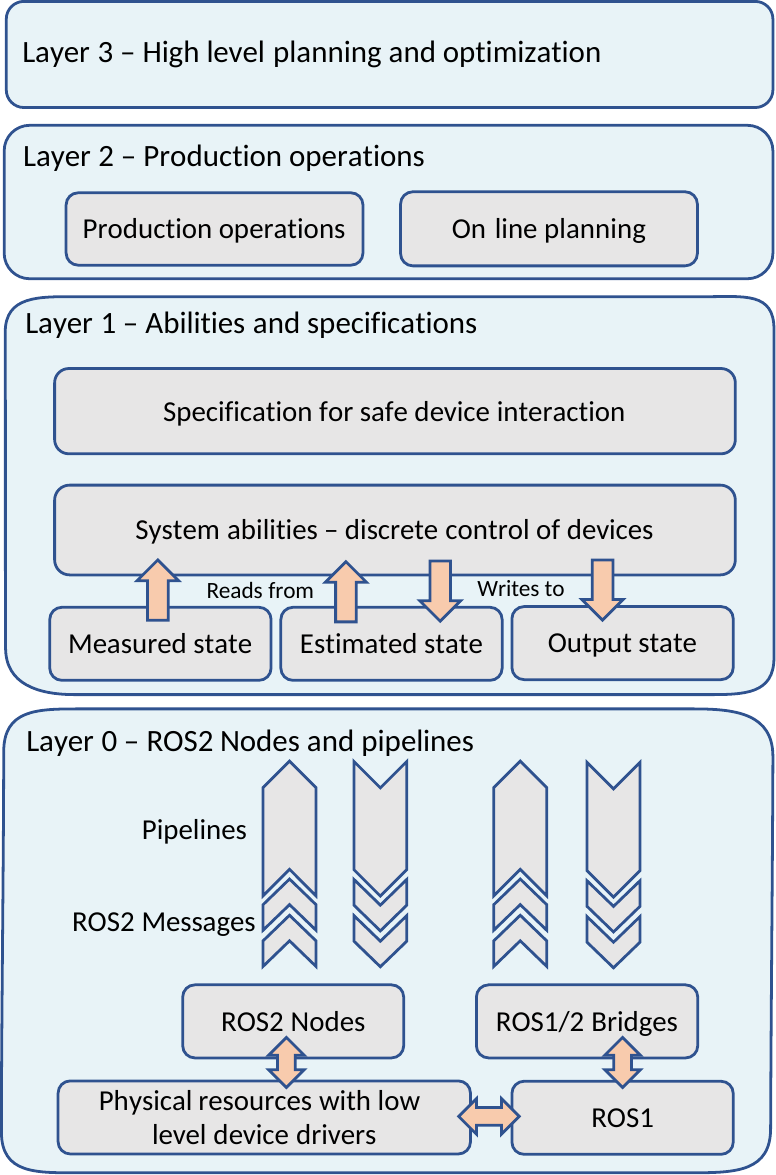}
  \caption{Layers in the proposed control architecture.}
  \label{fig:architecture}
\end{figure}

In this paper we describe an architecture for composing heterogeneous ROS2 nodes into a hierarchical automation system, an overview of which can be seen in Figure~\ref{fig:architecture}.
The automation system is divided into four layers. Layer 0, \emph{ROS2 Nodes and pipelines}, concerns the individual device drivers and ROS2 nodes in the system. As SP uses EFA which are based on transitions that update \emph{variables} as the core modeling formalism, \emph{transformation pipelines} are defined in layer 0 to map ROS2 messages coming from and going out to the nodes in the system to variables within SP.
State in SP is divided into \emph{measured state} - coming from the ROS2 network, \emph{estimated state} - inferred from previous actions of the control system, and \emph{output state} - to be sent out on the ROS2 network. Layer 1, \emph{Abilities and specifications}, concerns modeling the \emph{abilities} of the different resources, which are the low-level tasks that the resources can perform. Depending on the system state, abilities can be \emph{started} which trigger state changes of the \emph{output variables} which is eventually transformed by the pipelines in layer 0 to ROS2 messages. Abilities are modeled in two steps: first individually, then the system specific interactions are modeled as global specifications. Specifications in layer 1 is generally safety oriented: ensuring that nothing ``bad'' can happen. Layer 2, \emph{Production operations}, defines the production \emph{operations}~\cite{bengtsson2009origin} of the automation system. Production operations are generally defined on a high abstraction level (e.g. ``assemble part A and part B''). Production operations are dynamically matched to sequences of suitable abilities during run-time. Finally, layer 3, \emph{High level planning and optimization} represents a high level planning or optimization system that decides in which order to execute the production operations of layer 2. Layer 3 is not further discussed in this work.

\section{Layer 0 - ROS2 Nodes and pipelines}
\label{sec:layer0}
Layer 0 consist of the already given device drivers such as motor controllers or individual sensors. However, they can also be of more high level nature, consider for example a robot driver with a path planning algorithm. If the drivers are not already ROS2 nodes, they should be wrapped in thin layers for interfacing between the control system and the underlying device drivers. ROS2 nodes are as much as possible kept stateless to handle complicated state machines on the higher levels. Table~\ref{tab:setup} shows an overview of the ROS2 nodes in the use case described in Section~\ref{sec:use-case}.

\begin{table*}[h]
  \centering
  \begin{tabular}{l l l l l l l l} \toprule
    No. & Name & ROS v. & Computer & OS & Arch. & Network & Explanation \\ \toprule
    1 & Tool ECU & Bo+Me & Rasp. Pi & Ubuntu 18 & ARM & LAN1 & Smart tool and lifting system control \\
    2 & RSP ECU & Bo+Me & Rasp. Pi & Ubuntu 18 & ARM & LAN1 & Pneumatic conn. control and tool state \\
    3 & Dock ECU & Bo+Me & Rasp. Pi & Ubuntu 18 & ARM & LAN1 & State of docked end-effectors \\
    4 & MiRCOM & Bouncy & LP Alpha & Ubuntu 18 & amd64 & LAN2+VPN & ROS2 (VPN) to/from REST (LAN2) \\
    5 & MiR & Kinetic & Intel NUC & Ubuntu 16 & amd64 & LAN2 & Out-of-the-box MiR100 ROS Suite \\
    6 & RFIDCAM & Bouncy & Desktop & Win 10 & amd64 & LAN1 & Published RFID and Camera data \\
    7 & UR10 & Bo+Kin & Desktop & Ubuntu 16 & amd64 & LAN1 & UR10 ROS Suite \\
    8 & DECS & Bouncy & Laptop & Ubuntu 18 & amd64 & LAN1+VPN & Sequence Planner \\
    \bottomrule
  \end{tabular}
  \caption{Overview of the nodes in the use case.}
  \label{tab:setup}
\end{table*}

ROS2 has a much improved transport layer compared to ROS1, however, ROS1 is still far ahead of ROS2 when it comes to the number of packages and active developers. In order to embrace the strengths of both ROS1 and ROS2, i.e. to have an extensive set of developed robotics software (e.g. \emph{MoveIt!}~\cite{moveit}) and a robust way to communicate, ROS1 nodes are routinely bridged to the ROS2 network.

The fact that ROS uses typed messages means that the state of our low-level controller can be automatically inferred from the message types used by the involved ROS2 nodes. For simpler devices it is straight-forward to simply ``wire'' a topic into a set of \emph{measured state} variables in SP. However, this may not always be the case. To be able to support a wide variety of existing ROS2 nodes, we introduce a concept of applying \emph{pipelines} to transform the messages on the ROS2 network into SP state variables.

Pipelines are typed and can be composed in a graph-like manner to include merging and broadcasting to different endpoints, which allows graphical visualization of the different processing steps. The pipelines themselves are not part of the EFA model underpinning the control, but as they are typed and their processing logic is well isolated, it is straight forward to apply traditional testing methods (e.g. unit testing) to them to ensure their correctness. This enables us to map complex device state into (possibly reduced) control state, as well as having a standardized way of for example aggregating, discretizing, renaming, etc. In SP, the pipelines are implemented using Akka Streams~\cite{akka}. A common pattern is to add a ``ticking'' effect to the end of a pipeline, which will have the effect of automatically generating new ROS2 messages based on the current SP state at some specified interval.

Consider the node controlling the smart nutrunner, node number one in Table~\ref{tab:setup}. To control the node, a resource with state variables corresponding to the message structures in Listing~\ref{lst:smart-tool} is defined in SP.
Messages on the tool's state topic are mapped into \emph{measured state} and the \emph{output state} of SP is mapped to messages published on the tool's command topic. In this case, an automatic mapping of the messages to the SP state variables can be generated, which works by generating a pipeline transformation that matches the field names of the message type to SP variables of the correct type. To ease notation in the coming sections, a shorter variable name is introduced in the comments of Listing~\ref{lst:smart-tool}, where measured state is denoted with a subscript ``?'' and output state is denoted with a subscript ``!''.

\begin{listing}[ht]
\begin{minted}[mathescape,
               numbersep=5pt,
               frame=lines,
               fontsize=\footnotesize]{python}
# /smart_nutrunner/state
bool tool_is_idle                 # => $ti_?$
bool tool_is_running_forward      # => $tr_?$
bool programmed_torque_reached    # => $ttr_?$

# /smart_nutrunner/command
bool set_tool_idle                # => $ti_!$
bool run_tool_forward             # => $tr_!$
\end{minted}
\caption{Messages to and from the smart tool.}
\label{lst:smart-tool}
\end{listing}

\begin{figure}[ht]
  \centering
  \includegraphics[width=0.48\textwidth]{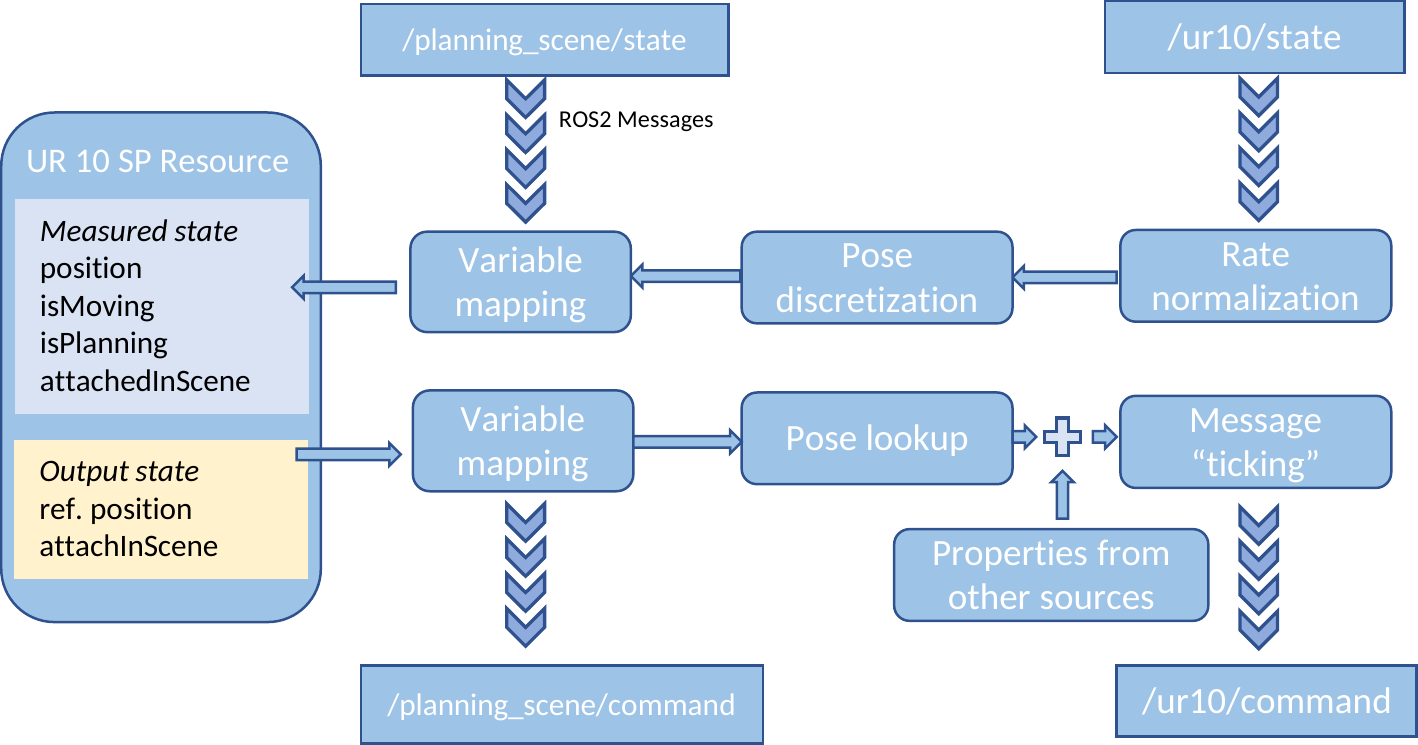}
  \caption{Schematic illustrating the pipelines for the topics to and from the UR10 node.}
  \label{fig:ur10pipeline}
\end{figure}

The state relating to the UR10 robot node (node 7 in Table~\ref{tab:setup}) is more complex than for the smart nutrunner. By applying the pipelines shown in Fig.~\ref{fig:ur10pipeline}, the robots position in space is discretized into an enumeration of named poses. On the output side, pipelines add information to the messages about whether the robot should plan its path, which planner it should use, and which type of move it should perform, etc. Message ticking and rate limiting steps are added as the last transformation steps in the respective pipelines to ensure a uniform update rate. Other properties which can be user-configured are merged later in the pipeline, allowing a way to manually override the messages generated by SP (for instance to lower the robot speed during testing). Note also that the state of the UR10 resource is collected from more than one topic.


\section{Layer 1 - Abilities and Specifications}
\label{sec:layer1}
Layer 1 forms the glue between the state captured by the transformation pipelines and the tasks the different resources in the system is able to perform. These tasks are defined in terms of the resource's state and are modeled as \emph{abilities}. Abilities are modeled per-resource and interactions between them are defined by \emph{specifications}.

\subsection{Abilities}
\label{sec:abilities}
An ability models a single task that a resource can perform. To track the state of an ability during execution, three boolean state variables are defined: \emph{isEnabled} (denoted $a^i$ for an ability $a$), \emph{isExecuting} ($a^e$), as well as \emph{isFinished} ($a^f$). A set of transitions define how an ability updates the system variables. ${ab}^\uparrow$ and ${ab}^\downarrow$ denotes the event corresponding to the transitions taken when starting and finishing the ability, respectively. Transitions to update the state of the ability, with the corresponding events ${ab}^{\rightarrow i}$, ${ab}^{\rightarrow e}$, ${ab}^{\rightarrow f}$, maps the resource state into the state of the ability, which allows the state of the ability to be synchronized with the \emph{measured state}. Each synchronization transition has a dual transition with the negation of guard of the original transition as its guard and an action that resets instead of sets the state variable.

In order to simulate an ability without its actual device, something that is required in order to perform the low level planning described in Section~\ref{sec:layer2}, an ability also need to define one or more \emph{starting} ($a^{s!!}$) and \emph{executing} ($a^{e!!}$) \emph{effects}. The effects are transitions which model how \emph{measured state} variables behave during the start and execution of an ability respectively.

\begin{figure}[ht]
  \centering
  \includegraphics[width=0.3\textwidth]{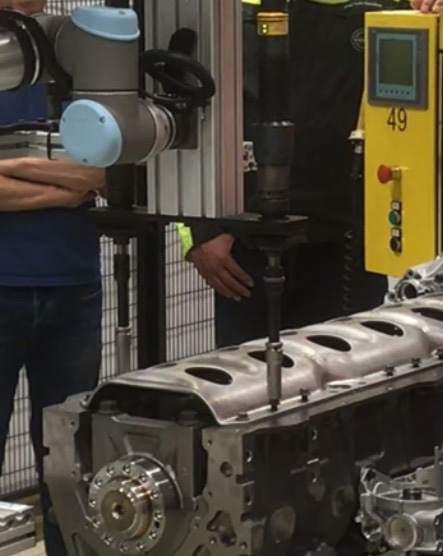}
  \caption{The smart nutrunner fastening the cover plate. Here operated by the UR10.}
  \label{fig:nutrunning}
\end{figure}

Consider again the smart nutrunner. The smart nutrunner is used to bolt down a cover plate onto the engine, as is shown in Figure~\ref{fig:nutrunning}. An ability, \var{runNut} ($a_{rn}$), that models the task of tightening a pair of bolts can be defined as follows. To be enabled, the tool should be in its \emph{isEnabled} state which can be reached by taking the transition ${ab}^{\rightarrow i} : \neg tf_? \land ti_? / a_{rn}^{i} := true$. When in this state, the start transition $a_{rn}^{\uparrow} : a_{rn}^{i} / tf_{!} := true  \land ti_{!} := false$ updates the \emph{output states} required to start the tool. Writing to these output state variables will, after passing through the pipeline transformation steps outlined in layer 0, eventually produce a message to the nutrunner topic on the ROS2 network. The transition ${ab}^{\rightarrow e} : tf_? \land \neg ti_? \land \neg ttr_? / a_{rn}^{e} := true$ synchronizes the executing state of the ability with the measured state, setting $a_{rn}^{e} := true$ if the tool is running forward and the pre-programmed torque has not yet been reached, as well as setting $a_{rn}^{e} := false$ when the this is not true.

The desired result of running the ability is to tighten a pair of bolts. As there are no sensors for keeping track of the bolts, the \emph{estimated} state variable $\hat{b} \in \text{(empty, placed, tightened)}$ is introduced. When the ability is executing and the programmed torque has been reached the tool should stop running forward, and $\hat{b}$ should be updated to 'tightened'. This is modeled in the transition $a_{rn}^{\downarrow} : a_{rn}^{e} \land ttr_? / ti_{!} := true \land tf_{!} := false \land \hat{b} = tightened$.

The effects of the ability are $a_{rn}^{s!!} : tf_{!} \land \neg ti_{!} / tf_{?} := true \land ti_{?} := false$, indicating that the tool is expected to start running forward after starting the ability and $a_{rn}^{e!!} : tf_? \land \neg ti_? / ttr_{?} := true$, indicating that during execution of the ability, the programmed torque is expected to eventually be reached.

For the UR10 a parameterized ability is defined: \var{moveToPosition}, $a_{um(p)}$ which takes a goal position ($u_p!$) as an input. As the node runs both the robot driver and the MoveIt! planning system, additional abilities are introduced: \var{attachInPlanningScene} and \var{detachInPlanningScene}, which sends the appropriate messages to MoveIt! for setting up the motion planning scene.

Additionally a \emph{restart} ability is introduced for the robot, \var{moveToPrevious}, which is enabled if the robot is not moving, its current position is an unknown position and its previous position is a known position. Restart abilities are only enabled during restart mode as described in Section~\ref{sec:restart}.

Finally, the human operator is modeled. The operator's role is to place bolts coming on the kitting AGV onto the cover plate. The human operator informs the system that he or she is finished with the task by acknowledging this on a smart watch. An ability for the operator, \var{placeBolt}, is defined that updates $\hat{b}$ to 'placed' during its finish transition.

\subsection{Specification}
The abilities defined so far, combined with the underlying transformation pipelines, can be used to run the system in an open loop fashion. As we have seen, the abilities are modeled on a per-resource basis, which allow individual testing of their behavior. However, the complexities of developing an automation system arise in the \emph{interaction} of the different resources.

In order to be able to work with individual devices, as well as isolating the complexities which arise from their different interactions SP relies heavily on modeling using global specifications. The abilities defined so far, together with a set of global specifications can be used to formulate a supervisor synthesis problem directly applicable to the EFA model. Using the method described in~\cite{miremadi2012bdd}, the solution to this synthesis problem can be obtained as additional \emph{guards} on the starting transition of the abilities. Examples of this modeling technique can be found in~\cite{bergagaard2015modeling,dahl2017automatic}. By keeping specifications as part of the model, there are fewer points of change which makes for faster and less error-prone development compared to changing the guard expressions manually.

For this case, a safety specification is added: when the robot is guiding the smart nutrunner to tighten a pair of bolts (see Figure~\ref{fig:nutrunning}), it is important that the \var{runNut} ability has started before moving down towards the cover plate, otherwise the tool will collide with the bolt. This can be modeled as the forbidden state specification
$u_p! = bp \land \hat{b} = \text{placed} \land \neg a_{rn}^e$, where $bp$ is the robot pose at the bolt location. It tells the system that it is forbidden for the robot to be at position $bp$ when an untightened pair of bolts is in place and the \var{runNut} ability has not been started.




\section{Layer 2 - Production operations}
\label{sec:layer2}
While layer 1 define all possibilities of what the system can safely do, layer 2 concerns making the system do something ``good''. For this use case it is to bolt the cover plate onto the engine. To achieve this, the high level production operation \var{TightenBoltPair} will be defined in this section.

In SP, production operations are modeled as \emph{goal states}. The goal states are defined as predicates over the system state. For the operation \var{TightenBoltPair}, the goal state is $\hat{b} = \text{tightened}$, i.e. the estimated bolt state should end up being 'tightened'. It also makes sense for the operations to have a precondition which ensures that the goal state is only activated when it makes sense. For this operation the precondition is $\hat{b} = \text{placed}$.

Modeling the production operations in this way does two things. First, it makes it possible to add and remove resources from the system more easily - as long as the desired goals can be reached, the upper layers of the control system does not need to be changed. Second, it makes it easier to model on a high level, eliminating the need to care about specific sequences of abilities.


Computing a plan for reaching a goal is done by finding a counter example using bounded model checking (BMC)~\cite{biere1999symbolic} on the EFA model of layer 1 (i.e. production operations can never start each other). Modern SAT-based model checkers are very efficient in finding counter examples, even for systems with hundreds of variables. BMC also has the useful property that counter examples have minimal length due to how the problem is unfolded into SAT problems iteratively. Additionally, well-known and powerful specification languages like \emph{Linear Temporal Logic} (LTL)~\cite{pnueli1977temporal} can be used. Being able to specify LTL properties means that operations can also contain local specification that should be active whenever the operation is executing. For example, reaching the state defined by the predicate $\phi_{good}$ while avoiding the state defined by the predicate $\phi_{bad}$ can be written as $\neg \phi_{bad} \Until \phi_{good}$. In the current implementation of SP, the SAT based bounded model checking capabilities of the nuXmv symbolic model checker~\cite{DBLP:conf/cav/CavadaCDGMMMRT14} is used, but planning engines based on PDDL~\cite{fox2003pddl2} could be plugged in as well.



\section{Control implementation}
\label{sec:implementation}
The system is executed synchronously based on the state of all connected resources. Operations execute when their preconditions are satisfied. The goals of each currently active operation are conjuncted to form a planning problem on the form $\bigwedge\limits_{i=1}^n \Eventually (o_ig)$ for the goal states $(o_ig)$ of the currently active operations $1$ to $n$, where $\Eventually$ is the LTL operator specifying that the predicate eventually becomes true. The result of the planning problem gives a start order of the system's abilities. Abilities are allowed to execute if their preconditions are satisfied and they are the first in the current start order. When abilities are started, they are popped from the start order and the next one may start if its preconditions are fulfilled. This greedy behavior enables multiple abilities to start executing in parallel, in contrast to purely sequential planning frameworks.

Reactivity is crucial in human/robot collaboration -- it should be possible for plans to change on short notice. By modeling reasonably small tasks for the operations in layer 2, planning can be fast enough to be performed continuously. Plans can then be followed in a receding horizon fashion, enabling quick reaction to changes in the environment.

The execution system keeps track of all the transitions in the automation system, defined by the abilities and the specifications in layer 1. For example, when an ability is started, the action of its starting transition is executed, updating the state of the SP variables. This state update triggers involved pipelines to assemble new ROS2 messages based on the defined transformations steps (for example generated variable mappings) in layer 0. At the end of the pipeline the new message is sent out on the ROS2 network for the different nodes to process.

\subsection{Restart situations}
\label{sec:restart}
During execution the system will invariably reach an error, or restart, situation. Given that it is not feasible to measure all state of the system, it is likely that the \emph{estimated state} will cause out of sync errors (e.g. a tightened bolt will end up in its initial ``empty'' state after a control system restart). To resynchronize the control system online an operator needs support from the system, for example, guiding the operator to a precalculated state from where restart is safe~\cite{bergagaard2013calculating}. SP employs a variety of ways in which to get back into a known state. One of the most important, however, is also the simplest one: keeping state machines out of the Level 0 nodes! It is never desirable to be forced to reset individual devices to get back to a known state.

Given this, it is not unlikely that restart errors can be solved by re-planning. For example if the robot went offline, it is probable that the planner can find a way for the operator to perform the tasks instead. If this fails, SP can enter a restart mode, where the automatic execution of operations is paused. In this mode, operations can be \emph{reset}, where instead of planning to reach the \emph{goal} state of the operation, the aim is to reach a state in which its precondition is satisfied. After a successful reset the operation can be started again. In this restart mode, it is possible to activate a number of \emph{restart abilities} during planning. A restart ability resets a subsystem back to a known state from which execution can resume (see \var{moveToPrevious} in Section~\ref{sec:abilities}). Lastly, it is up to an operator to bring the \emph{estimated state} of the system back into sync with reality, either by changing the physical world (e.g. putting a missing part into place), or by changing the system state to reflect the reality.

\section{Conclusion}
\label{sec:conclusion}
This paper introduced Sequence Planner (SP) as an architecture to model and control ROS2 based collaborative and intelligent automation systems. The control architecture has been implemented on an industrial assembly station. Practical experience during development of the control system for the described use case suggest that ROS2 does in fact enable larger scale industrial automation systems to be built on top of it.

The layered architecture allows reasoning about production operations independently of which combination of resources are used to perform them, but at the same time the low level approach taken to planning and control enables a structured approach for error handling on the level of individual subsystem state. Compared to more high level planning frameworks, this can make recovery after errors possible in more scenarios.

SP is under continuous development here~\cite{sp}. The hope is that it could become a tool used by a wider audience within the ROS community.

\bibliographystyle{IEEEtran}
\bibliography{IEEEabrv,references}

\end{document}